\documentclass[11pt,a4paper]{article}

\usepackage[T1]{fontenc}
\usepackage{lmodern}
\usepackage[margin=1in]{geometry}
\usepackage{microtype}
\usepackage{graphicx}
\usepackage{booktabs}
\usepackage{amsmath}
\usepackage{amssymb}
\usepackage[round,authoryear]{natbib}
\usepackage{enumitem}
\usepackage{float}
\usepackage{placeins}
\usepackage{caption}
\usepackage{xurl}
\usepackage[hidelinks]{hyperref}

\setlist{nosep,leftmargin=*}
\hypersetup{
  pdftitle={Representing Entity Importance in AI Knowledge Systems: A Dual-Signal Framework of Audience Evaluation and Structural Authority},
  pdfauthor={Shen Xu}
}

\title{Representing Entity Importance in AI Knowledge Systems:\\A Dual-Signal Framework of Audience Evaluation and Structural Authority}
\author{Shen Xu\\Independent Researcher}
\date{}

\begin{document}
\maketitle

\begin{abstract}
AI knowledge systems require representations of entity importance for retrieval, recommendation, evidence selection, and knowledge-intensive reasoning. Yet importance is often reduced to a single score derived from either human response or graph structure. Such compression may discard distinctions that matter when an AI system must choose among entities for different tasks. This study introduces an interpretable dual-signal representation in which each entity is characterized by an audience-evaluation dimension and a structural-authority dimension.

The framework is evaluated using movie entities as an empirical validation domain. IMDb non-commercial datasets provide a rating-based audience ranking, Wikidata supports entity alignment, and English Wikipedia hyperlinks form the knowledge network on which PageRank estimates structural authority. Experiments on 482 entities and 13,690 directed relationships reveal a statistically significant but weak association between the two dimensions (Spearman $\rho=0.2275$, $p<0.001$). Their overlap is only 10\% in the top 10 and 34\% in the top 100, while entity-level divergence occurs in both directions.

The results show that audience evaluation and structural authority are non-redundant signals and should not automatically be collapsed into a single scalar notion of importance. The contribution is not a new ranking algorithm or learned embedding, but a minimal knowledge-representation framework and an empirical test of its dimensional necessity. The findings support task-aware AI knowledge systems that preserve distinct importance signals before applying context-specific selection or aggregation.
\end{abstract}


\section{Introduction}

\subsection{Entity Importance as a Knowledge-Representation Problem}

Knowledge representation determines what information an intelligent system can make explicit, what inferences it can support, and which distinctions it preserves for subsequent reasoning \citep{davis1993}. For AI systems that retrieve, recommend, summarize, or reason over external knowledge, entities are not merely items in a database. They are representational units whose attributes, relations, provenance, and relative importance can affect downstream selection.

Entity importance is commonly encoded as a scalar ranking signal. Search engines, recommendation platforms, and online marketplaces use clicks, views, ratings, or relevance estimates to order documents, products, people, and other entities. Information-retrieval research provides the foundations for these mechanisms \citep{baeza2011}, while learning-to-rank research shows how observed user behavior can improve ordering decisions \citep{joachims2002}. Graph-based methods provide another family of signals by deriving importance from relational structure rather than direct user response \citep{brin1998,page1999}.

The representational question is whether these signals describe one underlying property or preserve different kinds of information. If audience evaluation and structural authority are effectively interchangeable, a single importance value may be sufficient. If they diverge systematically, compressing them into one score can remove information that an AI knowledge system may need for task-aware entity selection.

\subsection{Why a Single Importance Score May Be Insufficient}

Audience evaluation describes how participating users assess an entity under an explicit rating procedure. Structural authority describes the entity's position within a network of relationships. The first is grounded in evaluative preference; the second is grounded in relational structure. Neither is inherently superior, because they answer different questions.

An entity with a strong audience-evaluation rank may have limited structural authority, while an entity with a weaker audience rank may occupy a central position in a knowledge network. PageRank estimates importance through links from other important nodes \citep{page1999}; HITS distinguishes hubs from authorities \citep{kleinberg1999}; and centrality research provides multiple accounts of network position \citep{freeman1978}. These approaches demonstrate that relational importance can emerge independently of direct audience response.

This distinction matters for AI knowledge systems. A conversational or retrieval-augmented system may need a widely appreciated entity in one context, a structurally authoritative reference in another, or a combination of both. A representation that preserves only one signal cannot expose this distinction to the downstream selection process.

\subsection{A Dual-Signal Representation}

This study models entity importance as a two-dimensional representation rather than a single scalar. For an entity $e$, the framework preserves an audience-evaluation component and a structural-authority component:
\begin{equation}
\mathbf{I}(e)=\left[A(e),S(e)\right].
\end{equation}
Here, $A(e)$ represents audience evaluation and $S(e)$ represents authority derived from the entity's position in a knowledge network. The representation is intentionally minimal and interpretable. It is not a learned embedding and does not prescribe a universal weighting between the two components.

The empirical purpose of the study is to test whether retaining both components is necessary. Strong rank correlation and high agreement among top-ranked entities would suggest substantial redundancy. Weak agreement and large divergence would indicate that the two dimensions preserve different information.

\subsection{Research Questions and Contributions}

The study addresses three research questions:
\begin{itemize}
  \item \textbf{RQ1:} Are audience evaluation and structural authority sufficiently correlated to be treated as a single dimension of entity importance?
  \item \textbf{RQ2:} Do the two dimensions produce the same high-priority entity sets?
  \item \textbf{RQ3:} What representational information is exposed by entities whose positions diverge strongly across the two dimensions?
\end{itemize}

The study makes three contributions. First, it formalizes an interpretable dual-signal representation of entity importance for AI knowledge systems. Second, it provides a reproducible operationalization using IMDb ratings, Wikidata alignment, and Wikipedia hyperlink authority. Third, it empirically demonstrates that audience evaluation and structural authority are non-redundant in the selected domain, establishing a basis for preserving both signals before task-specific aggregation or decision making.

\section{Related Work}

\subsection{Knowledge Representation and Entity Modeling}

Knowledge representation is not only a choice of data format; it determines which distinctions an intelligent system can use in inference and action \citep{davis1993}. Knowledge graphs instantiate this principle by modeling entities and relations explicitly \citep{hogan2021}. Their relational structure supports entity linking, knowledge discovery, completion, reasoning, and knowledge-aware applications \citep{nickel2016,ji2022}.

Most knowledge-graph research focuses on representing entity attributes and relations, predicting missing links, or learning vector embeddings. Relative importance is often introduced later as a ranking or centrality score. This separation can obscure whether importance itself should be represented as a multidimensional property. The present study focuses on that narrower representational question.

\subsection{Audience Signals and Entity Ranking}

Behavioral and evaluative signals are widely used to prioritize entities. Clicks provide implicit evidence of attention, while ratings provide explicit evaluative preference. Click-through data can improve search ordering \citep{joachims2002}, but accumulated exposure can reinforce already prominent items and produce popularity bias \citep{abdollahpouri2019}. Ratings have a different limitation: they reflect evaluation among participating users rather than direct population-level attention.

Entity-oriented search treats entities as primary units of information access rather than incidental document mentions \citep{balog2018}. In such systems, audience-derived signals can be useful, but they do not encode the entity's relational role within a knowledge structure.

\subsection{Structural Authority in Knowledge Networks}

Web search established that structural information can complement textual and behavioral signals. The early Google architecture incorporated hyperlink analysis \citep{brin1998}, while PageRank formalized authority transfer through incoming links \citep{page1999}. HITS distinguishes hubs from authorities \citep{kleinberg1999}, and centrality research offers multiple measures of structural position \citep{freeman1978}.

These methods share a common premise: importance can emerge from relationships. However, graph-based authority is typically used as a ranking mechanism rather than explicitly retained alongside an independent audience-evaluation dimension. The degree to which the two signals are redundant is therefore an empirical and representational question.

\subsection{Knowledge-Augmented AI Systems}

Modern AI systems increasingly combine parametric models with external retrieval and structured knowledge. Retrieval-augmented generation demonstrates how retrieved evidence can support knowledge-intensive language tasks \citep{lewis2020}, while knowledge graphs provide explicit entity-relation structures for reasoning and semantic access \citep{hogan2021,ji2022}.

These systems must often choose which entities or evidence to surface. A single importance score may be convenient, but it can hide why an entity is prioritized. Preserving separate audience and authority components offers an interpretable representation from which a downstream system may make task-dependent choices. This study evaluates whether the two components contain sufficiently different information to justify that separation.

\section{Methodology}

\subsection{Research Design and Formal Representation}

This study adopts an empirical comparative design. It does not propose a new ranking algorithm, learned representation, or downstream AI architecture. Instead, it defines a minimal dual-signal representation and tests whether its two dimensions are empirically redundant.

Let $r_A(e)$ denote the audience-evaluation rank of entity $e$, let $r_S(e)$ denote its structural-authority rank, and let $N$ be the number of entities. For interpretability, each rank can be mapped to a normalized importance value in $[0,1]$:
\begin{align}
A(e) &= 1-\frac{r_A(e)-1}{N-1},\\
S(e) &= 1-\frac{r_S(e)-1}{N-1}.
\end{align}
Higher values indicate stronger importance on the corresponding dimension. The entity representation is then:
\begin{equation}
\mathbf{I}(e)=\left[A(e),S(e)\right].
\end{equation}
This transformation preserves the ordinal information used in the experiments. The framework does not assume that $A(e)$ and $S(e)$ should receive fixed weights. Instead, it preserves both dimensions so that a downstream AI system could apply task-specific selection or aggregation.

\begin{figure}[tbp]
  \centering
  \includegraphics[width=0.94\linewidth]{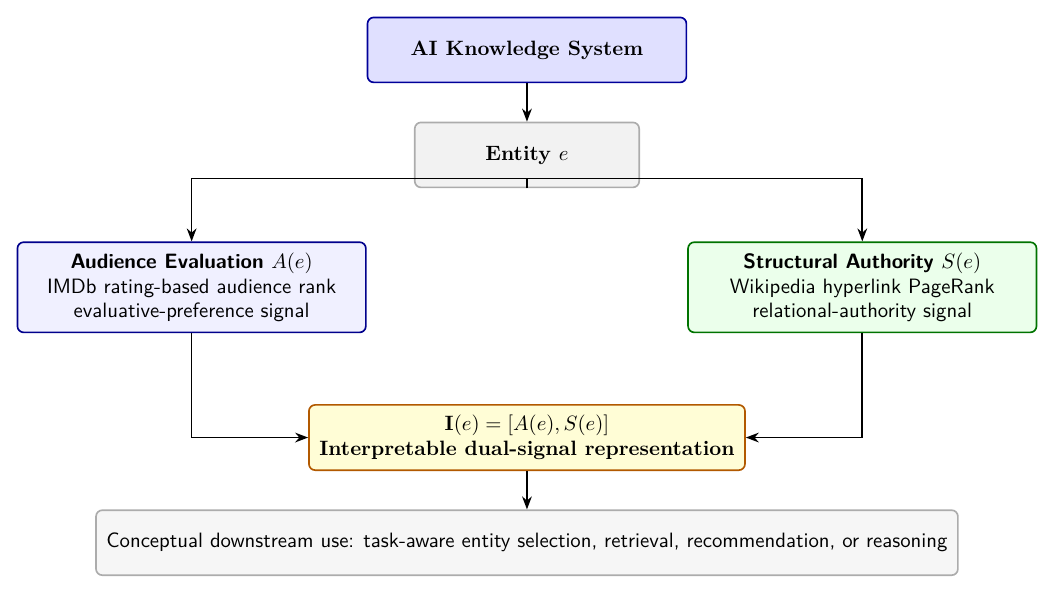}
  \caption{Dual-signal entity-importance representation for AI knowledge systems. Audience evaluation and structural authority are preserved as separate, interpretable dimensions before any task-specific downstream use.}
  \label{fig:framework}
\end{figure}

Figure~\ref{fig:framework} summarizes the framework. The empirical analysis tests dimensional redundancy: if audience evaluation and structural authority produce similar orderings and top-entity sets, separate representation adds little information; if they diverge, collapsing them into a single scalar would discard meaningful distinctions.

\subsection{Dataset Construction}

Movies are selected as an empirical validation domain because they combine measurable audience-evaluation signals with rich knowledge relationships. Movie entities are connected with directors, actors, genres, historical contexts, cultural references, and other entities. The domain therefore provides a controlled environment for testing the representational framework; movies are the validation setting rather than the theoretical research target.

The candidate set was constructed from the IMDb non-commercial \texttt{title.basics} and \texttt{title.ratings} datasets \citep{imdbdatasets}. The two tables were merged by the IMDb title identifier (\texttt{tconst}). Records were retained when they represented a movie, were marked as non-adult, had a known release year, had an average rating of at least 7.0, and had received at least 50,000 votes. Eligible records were ordered by average rating in descending order and then by vote count in descending order; the first 500 records formed the candidate set. Table~\ref{tab:selection} summarizes these rules.

\begin{table}[tbp]
  \centering
  \caption{IMDb candidate-set construction rules.}
  \label{tab:selection}
  \small
  \begin{tabular}{p{4.5cm}p{8.2cm}}
    \toprule
    Field or operation & Rule \\
    \midrule
    Title type & \texttt{titleType = movie} \\
    Adult-content flag & \texttt{isAdult = 0} \\
    Release year & Known value (not missing) \\
    Average rating & At least 7.0 \\
    Vote count & At least 50,000 \\
    Ordering & Average rating descending, then vote count descending \\
    Candidate-set size & First 500 eligible records \\
    \bottomrule
  \end{tabular}
\end{table}

Wikidata was used for cross-source entity alignment \citep{vrandecic2014}. Each candidate was queried by its IMDb identifier through Wikidata property P345, and the corresponding English Wikipedia article was retrieved when available. The final analytical sample contains 482 entities that had a usable cross-source mapping and participated in the induced Wikipedia hyperlink graph. Candidates without a complete mapping or without a retained intra-sample graph connection did not enter the final comparison. The reduction from 500 candidates to 482 analytical entities therefore reflects alignment and graph-construction coverage rather than an additional evaluation threshold.

The resulting dataset integrates three sources:
\begin{itemize}
  \item \textbf{IMDb non-commercial datasets:} provide identifiers, titles, release years, weighted average ratings, and vote counts used to construct the audience-evaluation dimension.
  \item \textbf{Wikidata identifiers:} align IMDb records with English Wikipedia pages.
  \item \textbf{English Wikipedia hyperlinks:} provide directed relationships among the aligned movie entities.
\end{itemize}

The derived datasets, analysis scripts, and reported results are archived in \href{https://github.com/ShenXuAkaEkstasis/entity-ranking-comparison/tree/9f14a19}{repository commit \texttt{9f14a19}}.

\subsection{Knowledge Graph Construction}

The knowledge graph is represented as a directed graph $G=(V,E)$, where $V$ is the set of movie entities and $E$ is the set of hyperlinks between their English Wikipedia pages. Hyperlinks were collected through the English Wikipedia MediaWiki Action API using the page-links module \citep{mediawikiapi}. Redirect resolution was enabled, continuation tokens were followed until all available links for a page had been retrieved, and URL-encoded page titles were normalized before matching.

The analysis uses an induced network over the aligned candidate entities: a hyperlink was retained only when both its source and target belonged to the aligned movie set. Self-links were removed, and duplicate source-target pairs were collapsed. Edge direction follows the original Wikipedia hyperlink direction. These rules make the graph a directed entity network rather than a general-purpose Wikipedia graph.

The resulting network contains 482 nodes and 13,690 directed edges. It has an average degree of 56.80, a density of 0.059049, and one weakly connected component. These properties permit PageRank to distribute authority across the full analytical entity set.

\subsection{Operationalizing the Two Dimensions}

\subsubsection{Audience Evaluation}

The audience-evaluation dimension is not IMDb's proprietary popularity chart. It is a rating-based audience ranking constructed from the public IMDb non-commercial datasets. Within the final 482-entity sample, entities are ordered by weighted average user rating in descending order. The source candidate list had already been ordered by rating and then by vote count, so vote count resolves equal-rating cases. Lower numerical rank indicates stronger audience evaluation under the study's explicit inclusion and tie-breaking rules.

This measure captures explicit evaluation among participating users, with vote count used for eligibility and tie breaking. It is not a direct measure of exposure, attention, or population-level popularity.

\subsubsection{Structural Authority}

PageRank is applied to the directed Wikipedia hyperlink network. For a node $X$, the score is defined as:
\begin{equation}
PR(X)=\frac{1-d}{N}+d\sum_{T_i\rightarrow X}\frac{PR(T_i)}{C(T_i)},
\end{equation}
where $d$ is the damping factor, $N$ is the number of entities, $T_i$ denotes an entity linking to $X$, and $C(T_i)$ is the number of outgoing links from $T_i$. The calculation uses NetworkX's PageRank implementation with $d=0.85$ (\texttt{alpha=0.85}) \citep{hagberg2008}; other numerical settings, including dangling-node handling, follow the NetworkX defaults. PageRank scores are sorted in descending order and converted into ordinal ranks, with lower numerical rank indicating greater structural authority.

\subsection{Testing Representational Redundancy}

The evaluation uses three complementary tests. First, Spearman rank correlation measures monotonic agreement between the complete audience-evaluation and structural-authority orderings. Second, Top-$K$ overlap tests whether the two dimensions produce the same high-priority entity sets at $K=10,25,50,$ and $100$. Third, entity-level divergence analysis identifies examples with large rank differences and shows what would be hidden if the dimensions were collapsed into a single value.

The tests do not estimate downstream AI performance. They evaluate a prerequisite representational question: whether the two signals preserve sufficiently different information to justify separate encoding.

\section{Experiments and Results}

\subsection{Dataset Statistics}

Table~\ref{tab:stats} summarizes the constructed network. The graph has one weakly connected component, meaning that every selected entity participates in the same underlying relational structure when edge direction is ignored. This is important because structural authority emerges from relationships rather than isolated entity attributes.

\begin{table}[tbp]
  \centering
  \caption{Knowledge-network statistics.}
  \label{tab:stats}
  \begin{tabular}{lr}
    \toprule
    Metric & Value \\
    \midrule
    Entities / nodes & 482 \\
    Directed edges & 13,690 \\
    Average degree & 56.80 \\
    Density & 0.059049 \\
    Weakly connected components & 1 \\
    \bottomrule
  \end{tabular}
\end{table}

\subsection{Dimensional Agreement: Rank Correlation}

The Spearman correlation between the audience-evaluation rank and structural-authority rank is:
\begin{equation}
\rho=0.2275, \qquad p<0.001.
\end{equation}
The exact calculation from the 482-entity dataset yields $\rho=0.2275223$ and $p=4.45\times10^{-7}$. The association is statistically significant but weak.

\begin{figure}[tbp]
  \centering
  \includegraphics[width=0.88\linewidth]{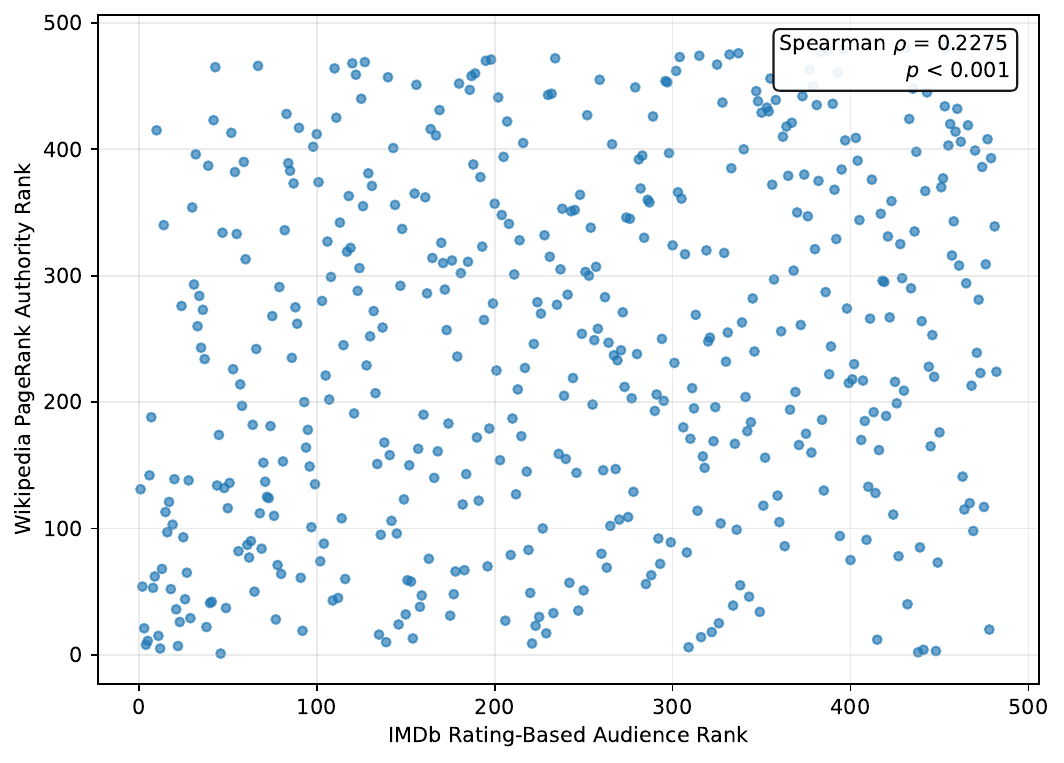}
  \caption{Agreement between the audience-evaluation and structural-authority dimensions. Each point represents one entity; lower numerical ranks indicate greater importance.}
  \label{fig:scatter}
\end{figure}

Figure~\ref{fig:scatter} shows the complete rank distribution. Each point represents one entity. Because lower numerical values indicate greater importance in both orderings, redundant dimensions would produce strong concentration around the diagonal. The broad dispersion instead indicates that entities with similar audience ranks can occupy very different authority ranks, and vice versa.

\subsection{\texorpdfstring{Decision-Set Agreement: Top-$K$ Overlap}{Decision-Set Agreement: Top-K Overlap}}

Correlation describes overall agreement but does not show whether the highest-priority entities are the same. Table~\ref{tab:topk} reports Top-$K$ overlap. Only one entity appears in both top-10 lists. The overlap rises with $K$, but remains 34\% even at the top 100.

\begin{table}[tbp]
  \centering
  \caption{Top-$K$ overlap between the audience-evaluation and structural-authority rankings.}
  \label{tab:topk}
  \begin{tabular}{lrr}
    \toprule
    Ranking scope & Shared entities & Overlap \\
    \midrule
    Top 10 & 1 of 10 & 10\% \\
    Top 25 & 6 of 25 & 24\% \\
    Top 50 & 15 of 50 & 30\% \\
    Top 100 & 34 of 100 & 34\% \\
    \bottomrule
  \end{tabular}
\end{table}

\begin{figure}[tbp]
  \centering
  \includegraphics[width=0.78\linewidth]{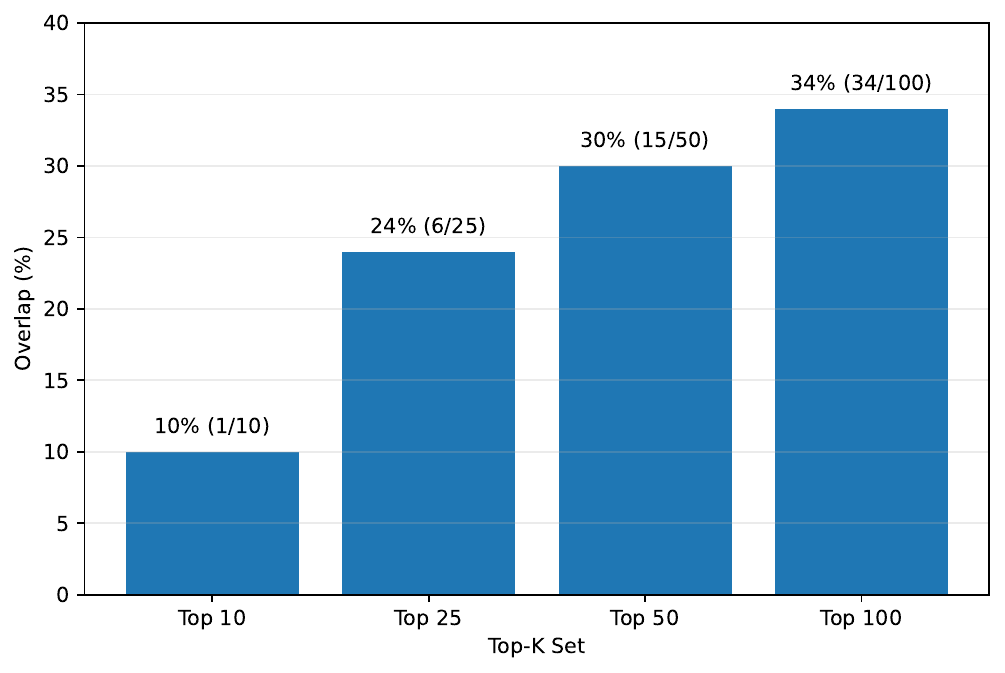}
  \caption{Top-$K$ overlap between the audience-evaluation and structural-authority rankings. Labels show the overlap percentage and number of shared entities.}
  \label{fig:topk}
\end{figure}

Figure~\ref{fig:topk} visualizes this pattern. The limited overlap demonstrates that the two dimensions do not merely reorder the same small group; they yield substantially different candidate sets for a system that selects only highly ranked entities.

\FloatBarrier

\subsection{Entity-Level Divergence}

Table~\ref{tab:divergence} presents examples with substantial rank differences. Positive gaps indicate stronger structural authority relative to audience evaluation, whereas negative gaps indicate stronger audience evaluation relative to structural authority.

\begin{table}[H]
  \centering
  \caption{Examples of divergence between the two representation dimensions. Lower ranks indicate greater importance. The gap is audience-evaluation rank minus structural-authority rank.}
  \label{tab:divergence}
  \small
  \begin{tabular}{p{5.4cm}rrr}
    \toprule
    Entity & Audience rank & Authority rank & Gap \\
    \midrule
    \multicolumn{4}{l}{\textit{Higher structural authority}} \\
    E.T. the Extra-Terrestrial & 438 & 2 & 436 \\
    Crouching Tiger, Hidden Dragon & 448 & 3 & 445 \\
    Avatar & 415 & 12 & 403 \\
    Titanic & 309 & 6 & 303 \\
    Slumdog Millionaire & 316 & 14 & 302 \\
    \addlinespace
    \multicolumn{4}{l}{\textit{Higher audience evaluation}} \\
    Fight Club & 10 & 415 & -405 \\
    Whiplash & 47 & 334 & -287 \\
    Seven Samurai & 31 & 293 & -262 \\
    12 Angry Men & 7 & 188 & -181 \\
    The Godfather Part II & 6 & 142 & -136 \\
    \bottomrule
  \end{tabular}
\end{table}

The first group contains entities whose network positions are much stronger than their audience-evaluation ranks. The second group contains entities with very strong audience ranks but substantially weaker PageRank positions. These cases should not be interpreted as judgments about intrinsic film quality. They demonstrate that the two dimensions preserve different information about the same entities and that scalar compression would conceal the direction of disagreement.

\section{Discussion}

\subsection{Evidence for a Non-Redundant Representation}

The experiments show that audience evaluation and structural authority should not be treated as interchangeable. Their weak correlation, low Top-$K$ overlap, and bidirectional divergence indicate that the two components of $\mathbf{I}(e)$ preserve distinct information.

This result does not establish that two dimensions are always sufficient, nor does it claim that PageRank is superior to audience evaluation. The IMDb signal identifies which entities receive stronger ratings under the study's inclusion rules. PageRank identifies which entities occupy influential positions within the selected relationship structure. The representational contribution is to retain this distinction rather than forcing one signal to stand for the other.

\subsection{Implications for AI Knowledge Systems}

AI knowledge systems often face task-dependent selection problems. A recommendation task may value audience evaluation, an evidence-tracing task may value structural authority, and a broad exploratory task may need both. Preserving the two dimensions enables a downstream system to make that choice explicitly.

For example, a task-aware system could define a contextual utility function:
\begin{equation}
U(e\mid q)=w_A(q)A(e)+w_S(q)S(e),
\end{equation}
where the weights depend on task or query context $q$. This equation is a design implication rather than an evaluated model in the present study. The important point is that such task-dependent aggregation remains possible only when the component signals are preserved before combination.

Separate dimensions also improve interpretability. A system can report whether an entity was prioritized because it was strongly evaluated by users, structurally authoritative in the knowledge graph, or both. This provenance is obscured when all evidence is compressed into an unexplained scalar score.

\subsection{Implications for AI-Mediated Entity Visibility}

The findings also clarify the relationship between entity visibility and entity representation. In traditional digital environments, visibility is commonly associated with traffic, engagement, or ranking position. In AI-mediated environments, an entity may additionally depend on explicit representation, relational connectivity, and suitability for a knowledge-intensive task.

An entity with strong audience evaluation but weak structural representation may be selected differently from an entity with broad knowledge-network authority. The present study does not establish a causal relationship between PageRank and visibility in a specific large language model. It identifies a plausible representational mechanism and a measurable distinction that future AI-visibility research can test directly.

\section{Limitations and Future Research}

This study has several limitations. First, the empirical evaluation is limited to movies. The domain combines audience ratings with rich knowledge relationships, but the degree of signal divergence may differ for products, organizations, scientific concepts, people, or geographic entities.

Second, the graph is constructed from English Wikipedia hyperlinks. Wikipedia reflects editorial, cultural, and linguistic choices, and hyperlinks represent only one type of relation. A graph based on Wikidata properties, citations, co-occurrence, or domain-specific relations could produce different authority signals.

Third, PageRank represents only one structural measure. HITS, degree centrality, betweenness, eigenvector centrality, and relation-aware graph models may capture other dimensions of authority. Comparing multiple graph measures would clarify whether the observed divergence is specific to PageRank or general to structural importance.

Fourth, the IMDb dimension is an operational measure of audience evaluation rather than direct popularity, exposure, or attention. It is determined by weighted average ratings, vote-count tie breaking, and explicit inclusion thresholds. Future work should compare this signal with page views, search volume, click behavior, and temporal engagement.

Fifth, the dual-signal representation is deliberately minimal. It is not a learned embedding, does not model uncertainty or provenance quality, and omits semantic relevance, temporal change, and task context. These dimensions can be added in future extensions.

Finally, the study tests representational non-redundancy rather than downstream AI performance. Future work should evaluate whether preserving and contextually weighting multiple importance signals improves entity selection, factual grounding, recommendation diversity, explanation quality, or knowledge-intensive reasoning.

\section{Conclusion}

This study reframes entity importance as a knowledge-representation problem for AI systems. It defines an interpretable dual-signal representation, $\mathbf{I}(e)=[A(e),S(e)]$, that preserves audience evaluation and structural authority as separate dimensions.

Using IMDb non-commercial datasets, Wikidata entity alignment, and English Wikipedia hyperlinks, the empirical analysis constructs a knowledge network of 482 movie entities and 13,690 directed relationships. The two dimensions are statistically related but weakly correlated, overlap by only 10\% in the top 10 and 34\% in the top 100, and produce large entity-level divergences in both directions.

These findings provide evidence against automatically collapsing audience evaluation and structural authority into a single scalar representation. For AI knowledge systems, preserving distinct signals enables task-aware aggregation, clearer explanation, and more explicit control over why an entity is selected. The framework is intentionally modest: it does not claim improved AI performance, but establishes an empirical basis for representing entity importance as multidimensional rather than unitary.

\section*{Data and Code Availability}

The derived datasets and analysis scripts used to produce the reported results are available in the \href{https://github.com/ShenXuAkaEkstasis/entity-ranking-comparison}{public repository}. The archived experiment state reported in this manuscript is \href{https://github.com/ShenXuAkaEkstasis/entity-ranking-comparison/tree/9f14a19}{commit \texttt{9f14a19}}.

\small
\bibliographystyle{plainnat}
\bibliography{references}

\end{document}